\documentclass[10pt,twocolumn,letterpaper]{article}

\usepackage{cvpr}              

\usepackage{graphicx}
\usepackage{amsmath}
\usepackage{amssymb}
\usepackage{booktabs}
\usepackage{cite}
\usepackage{amsmath,amssymb,amsfonts}
\usepackage{algorithmic}
\usepackage{graphicx}
\usepackage{textcomp}
\usepackage{color}
\usepackage{xcolor}
\usepackage{amsthm}
\usepackage{algorithm,algorithmic}
\usepackage{empheq}
\usepackage{ascmac}
\usepackage{framed}
\definecolor{shadecolor}{gray}{0.95}
\usepackage{listings}
\usepackage{gensymb}
\usepackage{here}
\usepackage{capt-of,etoolbox}
\usepackage{captdef}
\usepackage{caption}
\usepackage{multicol}
\usepackage{multirow}

\newcommand{\secref}[1]{Sec.\ref{#1}}
\newcommand{\figref}[1]{Fig.\ref{#1}}
\newcommand{\tabref}[1]{Tab.\ref{#1}}
\newcommand{\myeqref}[1]{Eq.(\ref{#1})}

\newcommand{\argmin}{\mathop{\mathrm{argmin}}\limits}

\newtheorem{theorem}{Theorem}

\newcommand{\nerf}{NeRF}
\newcommand{\modelname}{$\nabla_t$NeRF($S_t$)}
\newcommand{\algname}{TeGRA}
\newcommand{\dataname}{EvTrack}
\newcommand{\PM}{pose \& motion}
\newcommand{\PMCAP}{Pose \& motion}
\newcommand{\se}{\mathfrak{s}\mathfrak{e}}

\definecolor{codegreen}{rgb}{0,0.6,0}
\definecolor{codegray}{rgb}{0.5,0.5,0.5}
\definecolor{codepurple}{rgb}{0.58,0,0.82}
\definecolor{backcolour}{rgb}{0.95,0.95,0.92}
\lstdefinestyle{mystyle}{
  backgroundcolor=\color{backcolour}, commentstyle=\color{codegreen},
  keywordstyle=\color{magenta},
  numberstyle=\tiny\color{codegray},
  stringstyle=\color{codepurple},
  basicstyle=\ttfamily\footnotesize,
  columns=[1]{fullflexible},
  columns=fixed,
  basewidth=0.5em,
  breakatwhitespace=false,         
  breaklines=false,                 
  captionpos=b,                    
  keepspaces=true,                 
  numbers=left,                    
  numbersep=5pt,                  
  showspaces=false,                
  showstringspaces=false,
  showtabs=false,                  
  tabsize=1,
  xleftmargin=5mm,
}
\newcommand{\sparcity}{99.8\%}
\newcommand{\invsparcity}{0.2\%}

\usepackage[pagebackref,breaklinks,colorlinks]{hyperref}

\usepackage[capitalize]{cleveref}
\crefname{section}{Sec.}{Secs.}
\Crefname{section}{Section}{Sections}
\Crefname{table}{Table}{Tables}
\crefname{table}{Tab.}{Tabs.}

\begin{document}

\title{Event-based Camera Tracker by $\nabla_t$NeRF}

\author{Mana Masuda$^*$, Yusuke Sekikawa$^\dagger$, Hideo Saito$^*$\\
$^*$Keio University, $^\dagger$Denso IT Laboratory\\
{\tt\small \{mana.smile, hs\}@keio.jp, ysekikawa@mail.d-itlab.co.jp}
}
\maketitle

\begin{abstract}
When a camera travels across a 3D world, only a fraction of pixel value changes; an event-based camera observes the change as sparse events.
How can we utilize sparse events for efficient recovery of the camera pose?
We show that we can recover the camera pose by minimizing the error between sparse events and the temporal gradient of the scene represented as a neural radiance field (\nerf).
To enable the computation of the temporal gradient of the scene, we augment NeRF's camera pose as a time function.
When the input pose to the NeRF coincides with the actual pose, the output of the temporal gradient of NeRF equals the observed intensity changes on the event's points.
Using this principle, we propose an event-based camera pose tracking framework called TeGRA which realizes the pose update by using the sparse event's observation.
To the best of our knowledge, this is the first camera pose estimation algorithm using the scene's implicit representation and the sparse intensity change from events.
\end{abstract}

\section{Introduction}
\label{sec:intro}
\begin{figure*}[tb]
    \centering
    \includegraphics[width=1.0\linewidth]{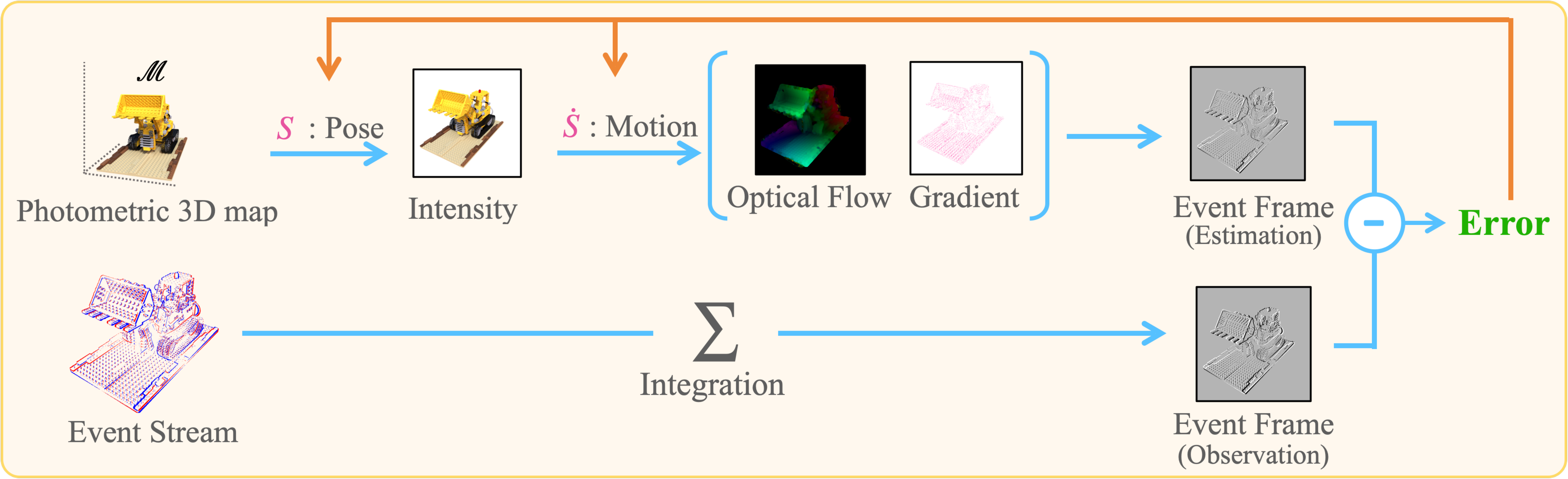}
    \figcaption{Tracking using event stream and explicit scene}
    \vspace{3mm}
    \label{fig:dense_method}
    \includegraphics[width=1.0\linewidth]{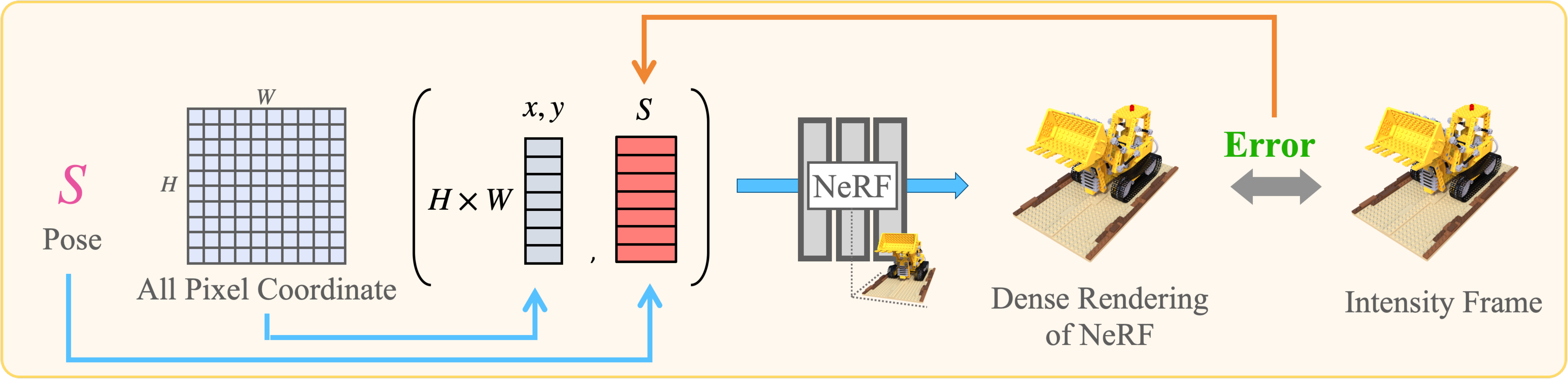}
    \figcaption{Tracking  using intensity frame and implicit scene}
    \label{fig:rgb-based}
    \vspace{3mm}
    \includegraphics[width=1.0\linewidth]{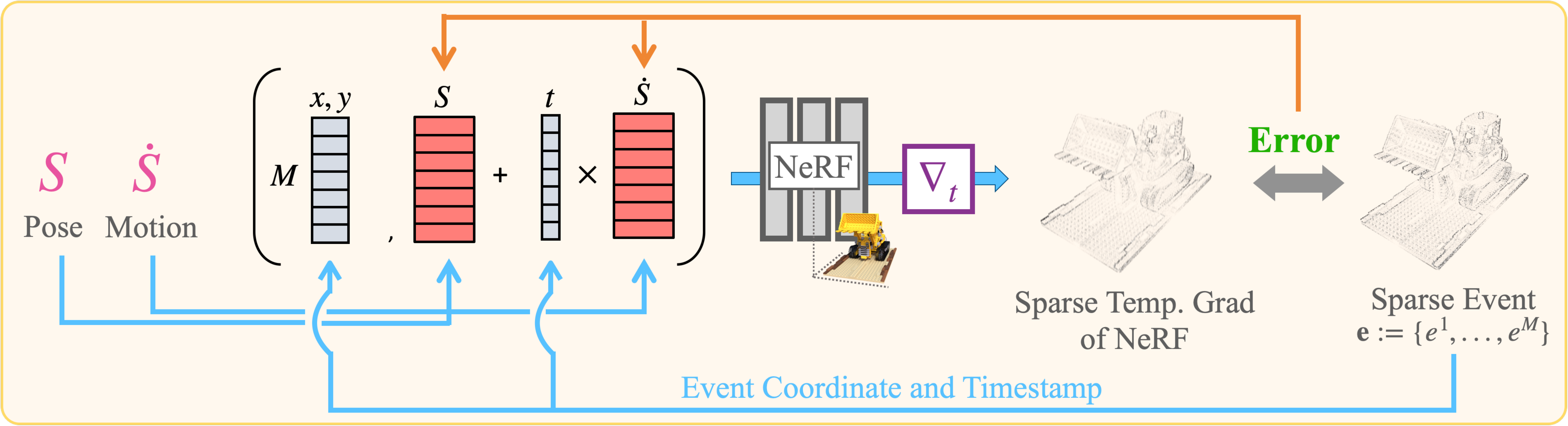}
    \figcaption{Tracking using event stream and implicit scene (\algname)}
    \label{fig:tegn}
    \vspace{3mm}
\end{figure*}

Camera localization/tracking is one of the fundamental functionality of computer vision.
The field of use lies in many applications like automotive, augmented reality, and robotics.
Event-based cameras detect sparse intensity changes with extremely high temporal resolution (\textgreater $10,000$ fps).
This unique feature makes it a suitable sensor for tracking fast-moving scenes, and many researchers have been exploring several approaches to utilize high-speed observation.
Recently, \cite{gehrig2018asynchronous,bryner2019event} showed the \PM\ is recovered by minimizing the error between the estimated intensity change and integrated events (\figref{fig:dense_method}).
Thanks to the low-latency nature of events, their method works well even in a rapid camera motion.
However, they need a dense operation to differentiate the error w.r.t \PM; it could not take advantage of the \textit{sparsity} of events.
Therefore, the computational cost increases linearly with the processing rate, making it difficult to run the algorithm in real-time on devices with limited computational resources.
\begin{figure*}
    \centering
    \includegraphics[width=0.95\linewidth]{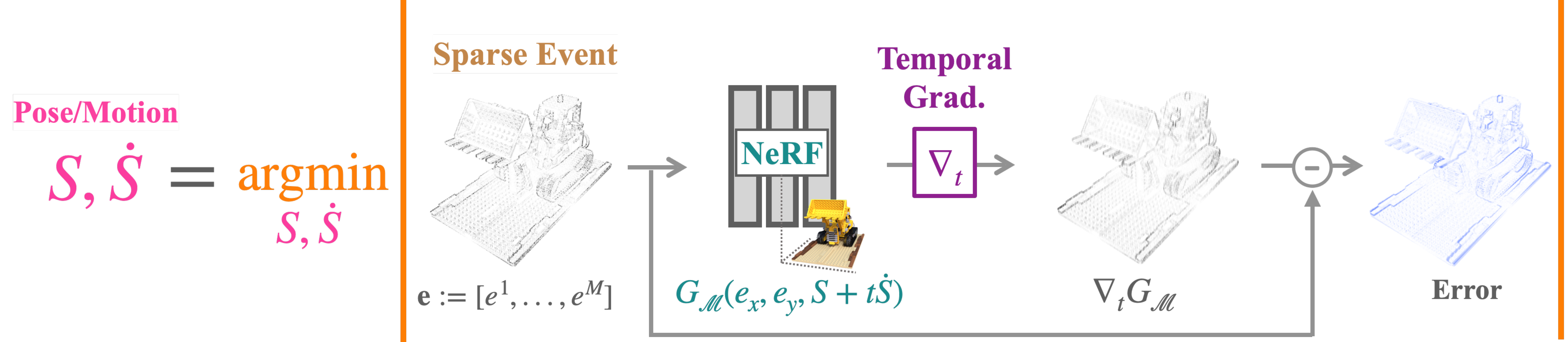}
    \vspace{4mm}
    \caption{{\color{orange}{$\operatorname{argmin}$}} $|${\color{violet}Temp. Grad.} of {\color{teal} NeRF($S_t$)} $-$ {\color{brown} Event}$|$ is {\color{magenta} Pose \& Motion}.}
    \label{fig:theorem}
    \vspace{-3mm}
\end{figure*}

\nerf\ \cite{mildenhall2020nerf} is an implicit light-field representation using a neural network, which enables unified perceptions of the 3D world that is difficult for the existing explicit MAP (e.g., CAD model).
Many \nerf-extention have been proposed \cite{tancik2022blocknerf} to model a complex 3D scene; we believe these advancements make the \nerf\ representation a novel candidate for representing 3D MAP for camera localization/tracking.
Utilization of \nerf\ as a 3D MAP for camera pose localization has already been explored \cite{yen2020inerf, lin2021barf}.
Their basic idea for realizing camera pose estimation is minimizing the error between the estimated intensity frame from \nerf\ and the observed intensity frame, w.r.t input camera pose (\figref{fig:rgb-based}).
However, these methods can not estimate the camera pose by using sparse intensity change.

These existing studies motivate one question; how can we utilize sparse events to recover the camera pose without converting them into dense frames?
We show:\\
\vspace{-4.5mm}
\begin{screen}
\vspace{-2mm}
\begin{center}
    {\small Pose is recovered by minimizing $|$\modelname\ - event$|$}.
\end{center}
\vspace{-6.5mm}
\end{screen}
When coordinates of events (where the event camera detects intensity changes) and the current camera pose estimate on the event's time $S_t$ are input to \nerf, it outputs the intensities at that point and time.
By viewing the input camera pose to \nerf\ as a function of time, we found that the temporal gradient of the intensity w.r.t the event's timestamp is the estimation of intensity changes on that pixel.
\PMCAP\ is obtained by minimizing $|$estimated intensity changes $-$ observed intensity changes (events)$|$ (Theorem-\ref{prop}, \figref{fig:theorem}).

Based on this principle, we propose an event-based camera pose tracking framework called minimization of the \modelname\ (\algname).
Unlike the conventional dense approach (frame-based), which operates in intensity-space, \algname\ works sparsely (event-based) in gradient-space; it updates the \PM\ by evaluating the error only for the pixels where the events have been observed.
By the sparse mechanism, the number of pixels to be evaluated is \textcolor{red}{\sparcity\ less} than the conventional dense algorithm on our created event dataset.
To the best of our knowledge, this is the first approach to realizing camera pose estimation using the implicit representation of the 3D scene and the sparse observation of intensity changes.
We provide the theoretical proof of \algname.
Furthermore, we created a photo-realistic event dataset for 6DoF camera pose tracking with a ground-truth pose called \dataname~(EVent-based TRACKing dataset). 
Using the \dataname, we experimentally proved the concept.

\section{Related Work}
\subsection{Camera Pose Tracking from Intensity Frame}
There is a long history in the field of camera pose tracking.
Many algorithms are based on frame-based observation, namely video sequence.
Methods based on KLT \cite{10.5555/1623264.1623280,Baker2004LucasKanade2Y} are among the most popular visual tracking algorithms.
KLT computes the camera pose by aligning the observed intensity frame with the known scene's intensity map using an image gradient.
Recently, the deep neural network (DNN) based feature extractor has been utilized to exploit more rich features \cite{chang2017clkn} than the raw pixel value.
This KLT-based method works well in many scenarios.
However, it easily collapses in a fast-moving scene, which induces a significant difference between the observed frame and the estimated frame. 
The difference makes the gradient-based algorithms trapped into a local minimum.
Another line of research utilizes DNN to directly regress the pose between pairs of images \cite{bertinetto2016staple,kart2019object,bertinetto2016fully,li2018high,voigtlaender2020siam}.
They have an advantage in computational efficiency (over the iterative gradient-based algorithm) because they can update the pose by a single forward pass of the network.
They may also suffer from performance degeneration when there is a significant difference in the pairs of images due to the fast camera motion.

Either gradient-based or regression-based, the problem due to the fast camera motion might be mitigated by using an expensive high-speed camera.
However, processing them at a higher rate is infeasible due to increased computational complexity.

\subsection{Camera Pose Tracking from Events}
An event-based camera \cite{chen2019live,eventcamera,prophesee,suh20201280,serrano2013128} is a bio-inspired vision sensor that mimics biological retinas.
It differs from a frame-based camera in its H/W design; its report per-pixel intensity changes as asynchronous event streams.
Thanks to this unique operation principle, event-based cameras have significant advantages over conventional frame-based cameras, e.g., low latency, high dynamic range (HDR), and blur-free.
The most important feature of the event-based camera for camera pose tracking would be its high temporal resolution.
Their temporal resolution is equivalent to \textgreater $10,000$ fps, making it a suitable sensor for robust tracking in fast-moving scenes.

\noindent\textbf{Use of Event frame for Camera Pose Estimation.}
There have been many attempts to utilize the distinct features of the event-based camera for tracking \cite{rebecq2017real,zihao2017event,gallego2017event,gehrig2018asynchronous,bryner2019event,alzugaray:3DV19,alzugaray2020haste,chamorro2020high,gehrig2020eklt}. 
Recently, KLT has been extended to event data to realize robust camera pose tracking in high-speed and HDR scenarios \cite{bryner2019event,gehrig2018asynchronous,gehrig2020eklt}.
They update camera pose and motion by minimizing the error in the integrated event frame between estimation and observation (\figref{fig:dense_method}).
The estimation is computed from the current estimate of the \PM\ using the pre-build 3D intensity map.
By utilizing the low-latency nature of events, their methods work well even in rapid camera motion.
However, computing the derivatives of the error w.r.t pose requires computationally intensive dense rendering of the 3D map.
The dense computation needs to be repeated until convergence for each frame.
Therefore, running the algorithm in real-time on devices with limited computational resources is difficult.
Our goal in this study is to realize an efficient sparse algorithm for camera pose tracking using the implicit representation of a 3D scene.

\subsection{Implicit Scene Representations}
Representing data, such as image, video, and shape using implicit neural representation gained much attention such as for data compression \cite{chen2019learning,mescheder2019occupancy}, novel-view synthesis \cite{mildenhall2019llff,mildenhall2020nerf,niemeyer2021giraffe}, 3D-shape modeling \cite{park2019deepsdf,chabra2020deep,atzmon2020sal,sitzmann2020implicit}, and image registration \cite{yen2020inerf,lin2021barf} to name a few.
\nerf\ \cite{mildenhall2020nerf} utilizes the implicit neural representations to represent the 4D light-field.
Given a set of images paired with a camera pose, \nerf\ learns the intensities of each pixel for a given camera pose.
Each pixel's intensity is computed by integrating the RGB color and density along the corresponding ray using the volumetric rendering technique.
Due to its flexibility, \nerf\ extensions are exploding beyond its original application of novel-view synthesis; e.g., 3D-shape reconstruction \cite{oechsle2021unisurf,wang2021neus}, disentangle lighting \cite{srinivasan2021nerv,boss2021nerd,Wizadwongsa2021NeX,zhang2021nerfactor}, image editing \cite{liu2021editing,zhang2021stnerf,yang2021objectnerf}, object separation \cite{xie2021fig,baatz2021nerf}, semantic label propagation \cite{Zhi:etal:ICCV2021}, modeling time-varying objects \cite{pumarola2021d,park2021nerfies,gafni2021dynamic,kwon2021neural,liu2021neural,li2021neural}, and depth estimation \cite{IchnowskiAvigal2021DexNeRF,wei2021nerfingmvs}.

\noindent\textbf{Use of \nerf\ for Camera Pose Estimation}
\label{sssec:nerf_track}
These novel functionalities realized by \nerf\ such as modeling lighting and moving objects would realize a unifying perception of the 3D world that is difficult for the existing explicit MAP (e.g., CAD model).
Furthermore, recent progress of \nerf-extention enabled it to model a large environment, such as a large house \cite{devries2021unconstrained}, and an entire city \cite{tancik2022blocknerf}.
We believe these advancements make the \nerf\ representation a novel candidate for representing 3D MAP for camera localization/tracking.
iNeRF \cite{yen2020inerf} uses the inverse of \nerf\ to estimate the camera pose.
BARF \cite{lin2021barf}, NeRF- \cite{wang2021nerf}, iMAP \cite{sucar2021imap}, NICE-SLAM\cite{zhu2021nice} realized simultaneous camera pose estimation and 3D scene reconstruction.
Their basic idea for realizing camera pose estimation is minimizing the error between the estimated intensity frame from \nerf\ and the observed intensity frame, w.r.t input camera pose (\figref{fig:rgb-based}).
These methods can not estimate the camera pose by using sparse intensity change.
We aim to derive a camera pose tracking algorithm using sparse intensity changes observation (i.e., events) to realize efficient camera pose tracking.

\section{Method}

\subsection{Preliminaries}
\label{sec:preliminaries}

\noindent\textbf{Problem Statement}
Our goal in this study is to develop an efficient camera pose tracking algorithm \algname\ using sparse observation of intensity-change event (IC-event) stream $\mathbf{e}_t$ as follows:
\begin{equation}
    \mathrm{TeGRA}: (\mathbf{e}_t, S^{\mathrm{ini}}_t, \dot{S}^{\mathrm{ini}}_t) \mapsto (S^{\mathrm{opt}}_t, \dot{S}^{\mathrm{opt}}_t),
\end{equation}
where $(S^{\mathrm{ini}}_t,\dot{S}^{\mathrm{ini}}_t)$ and $(S^{\mathrm{opt}}_t,\dot{S}^{\mathrm{opt}}_t)$ are the initial and optimized \PM\ at time $t$ respectively.
\algname\ utilizes a differentiable implicit representation of the static 3D world.

\noindent\textbf{NeRF}
This study assumes that the 3D scene is implicitly represented by \nerf\ \cite{mildenhall2020nerf}.
\nerf, $G_\mathcal{M}$ was initially proposed for novel-view synthesis; it represents the scene using a neural network that is differentiable w.r.t to its input pose ${S}$. 
The parameter of the \nerf, $\mathcal{M}$, is pre-trained before tracking. 
\nerf\ $G_\mathcal{M}$ takes an image coordinate $(x, y)$ and 6DoF pose $S\in \se (3)$ of the camera as inputs and renders the RGB intensity $\mathbf{c}$ at that coordinate;
\begin{equation}
\label{eq:nerf}
    G_\mathcal{M}(x, y, S)=\mathbf{c}.
\end{equation}

\noindent\textbf{IC-Event}
The IC-event stream $\mathbf{e}_t$ is a set of events observed in time interval $[t, t+\tau]$ as follows:
\begin{equation}
    \mathbf{e}_t = [e^1,...,  e^i, ..., e^M];\ \ e^i = [e_x^i, e_y^i, e_u^i, e_r^i]^{\mathrm{T}},
\end{equation}
where $(e_x^i, e_y^i)\in\mathbb{R}^2$ is the image coordinates where the intensity change has been detected, $e_u^i\in\mathbb{R}$ is the timestamp of the change, and $e_r^i\in\mathbb{R}$ is the intensity change value.
The intensity change value $e_r(x,y)$  on $(x,y)$ within time interval $\Delta t$ is defined using \textit{true} intensity $L(x, y, t)$ as follows:
\begin{equation}
\label{eq:def_event_rate}
    e_r(x,y) =\frac{L(x, y, t+\Delta t) - L(x,y, t)}{\Delta t}.
\end{equation}
Events are detected where the intensity change $e_r(x,y)$ exceeds the predefined threshold $\delta$.
For ease of discussion, we define $\bar{e}_u^i$, which is relative timestamp w.r.t time $t$; $\bar{e}_u^i:=e_u^i-t$.
That is $\bar{e}_u^i$ changes depending on the time $t$ we consider.
Some event-based cameras directly detect the intensity change $e_r$, such as Celex-V \cite{chen2019live}, or IC-event is obtained from high-speed video data.
We left the evaluation using the binary event for future work\footnote{At this time,
we consider there are two possible approaches to make \algname\ compatible with the binary-event; 1) Modify the loss function of \eqref{eq:theorem} to adapt to the binary-event (\secref{sssec:async_update}), 2) Convert the binary-event to IC-event using the timestamp (supplement-\ref{suppsec:b2e}). In this work, we use IC-event for simplicity and left the exploration for future work.
}.

\begin{figure*}[tb]
    \centering
    \includegraphics[width=0.9\linewidth]{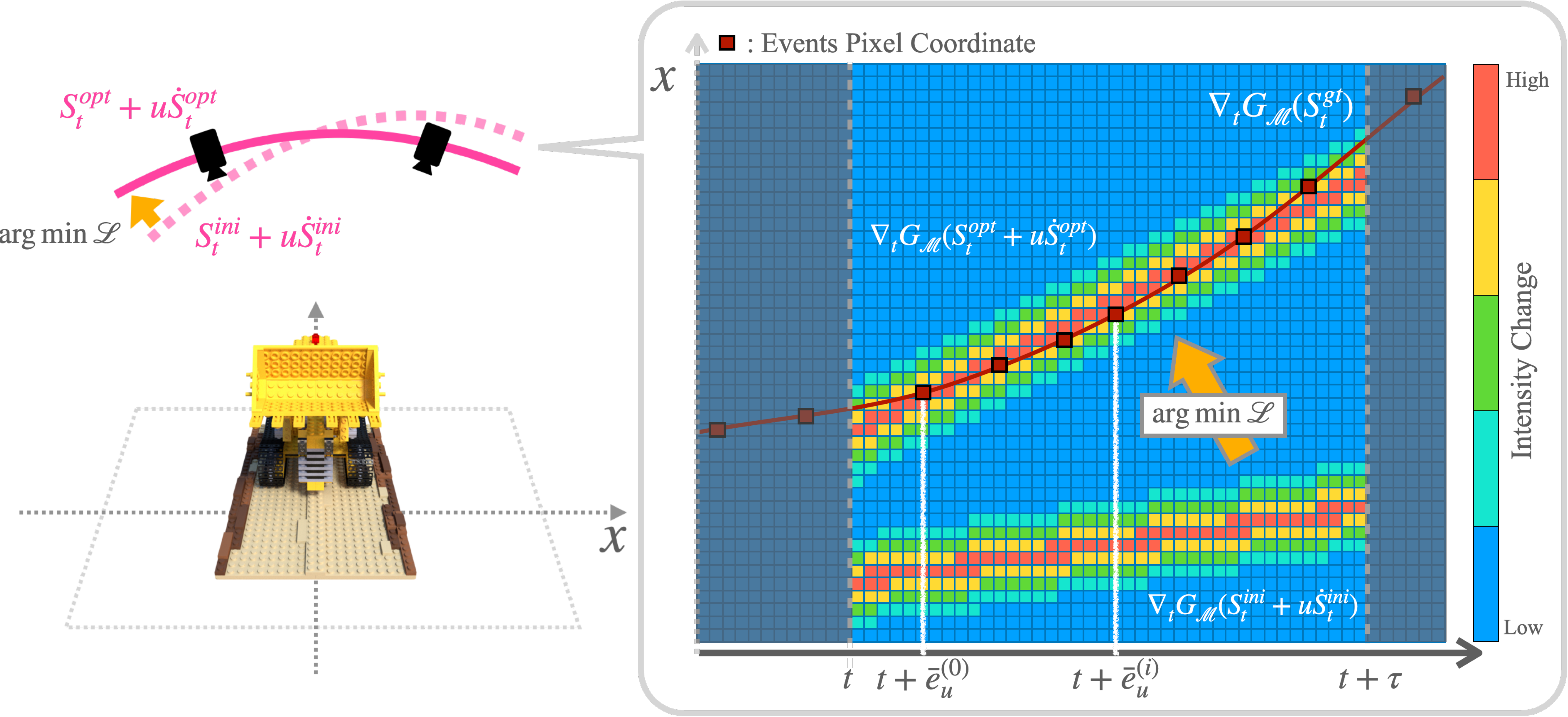}
    \caption{\PMCAP\ estimation (visual explanation of theorem-\ref{prop})}
    \label{fig:optimization}
\end{figure*}

\subsection{The answer of \modelname = 0 is true pose \& motion}
\label{sec:theorem}
By viewing the input camera pose to \nerf\ as a function of time, we found that camera \PM\ $(S,\dot{S})$ can be recovered by minimizing the error between the temporal gradient of the 3D-scene represented as \nerf\ $G_\mathcal{M}$ and IC-event stream $\mathbf{e}$ (\figref{fig:optimization}: visual explanation).
\begin{screen}
\begin{theorem}
\label{prop}
The minimizer of $| \nabla_t \mathrm{NeRF}(S_t) - \mathrm{event}|$ is \textit{true} pose $\&$ motion ($S^{\mathrm{gt}}_t, \dot{S}^{\mathrm{gt}}_t$):

\begin{equation}
\label{eq:theorem}
\begin{split}
    &S^{\mathrm{gt}}_t, \dot{S}^{\mathrm{gt}}_t \\
    &= \argmin_{S, \dot{S}}\underbrace{\sum_{i}\left\|\frac{\partial G_{\mathcal{M}}\left(e_x^i, e_y^i, S_t+\bar{e}_u^i\dot{S}_t\right)}{\partial \bar{e}_u^i}  - e_r^i\right\|_2}_{:=\mathcal{L}}.
\end{split}
\end{equation}
\end{theorem}
\end{screen}

\begin{proof}
To prove the theorem-\ref{prop}, we'll show the $i$-th element of the \myeqref{eq:theorem} equals to zero when we have the true \PM\ $({S}_t^{\mathrm{gt}},{\dot{S}}_t^{\mathrm{gt}})$.
Now, consider the temporal gradient of the \nerf\ of \myeqref{eq:nerf} for $i$-th event on time $e_u^{i}$:
\begin{equation}
\label{eq:nerf_grad}
\begin{split}
    &\frac{\partial G(e_x^i, e_y^i, {S}_t+\bar{e}_u^i\dot{{S}}_t)}{\partial \bar{e}_u^i}\\
    &=\lim_{\Delta u\rightarrow0} \frac{G(e_x^i, e_y^i, S_t+(\bar{e}_u^i+\Delta u)\dot{S}_t)-G(e_x^i, e_y^i, S_t+ \bar{e}_u^i\dot{S}_t)}{\Delta u}
\end{split}
\end{equation}
From the defenition of \nerf\ of \myeqref{eq:nerf}, $L(x, y, t)$ equals to the \nerf's output, when the estimated \PM\ is \textit{true}:
\begin{equation}
\label{eq:intensity_vs_nerf}
    L(x, y, t+\bar{e}_u^i) = G(x, y, S_t^{\mathrm{gt}}+\bar{e}_u^i\dot{{S}}_t^{\mathrm{gt}}).
\end{equation}
Pluging this relation to \myeqref{eq:nerf_grad}
\begin{equation}
\begin{split}
    &\frac{\partial G(e_x^i, e_y^i, S_t^{\mathrm{gt}}+\bar{e}^i\dot{S}_t^{\mathrm{gt}})}{\partial e_u^i} \\
    &= \lim_{\Delta u\rightarrow0}\frac{L(e_x^i, e_y^i, t+\bar{e}_u^i+\Delta u) - L(e_x^i, e_y^i, t+\bar{e}^i)}{\Delta u}\\
    &= \lim_{\Delta u\rightarrow0}\frac{L(e_x^i, e_y^i, e_u^i+\Delta u) - L(e_x^i, e_y^i, e_u^i)}{\Delta u}\\
\end{split}
\end{equation}
This equals to the definition of $e_r^i$ of \myeqref{eq:def_event_rate} when $\Delta t$ is sufficiently small.
\end{proof}

\subsection{Sparse Tracking Algorithm: TeGRA}
\label{sec:tracking_alg}
Using the theorem-\ref{prop}, we propose an event-based camera pose-tracking algorithm called TeGRA.
The algorithm minimizes $\mathcal{L}$ in \myeqref{eq:theorem} using gradient-decent (Fig. \ref{fig:tegn}).
The input to \algname\ is the IC-event stream $\mathbf{e}_t$ at time $t$ and the initial estimate of pose \& motion ($S^{ini}_t, \dot{S}^{ini}_t$). 
To obtain the derivative to update \PM, we differentiate the loss $\mathcal{L}$ w.r.t $(S_t, \dot{S}_t)$ through the temporal gradient of $G_\mathcal{M}$ on each event's timestamp $e_u^i$.
Then, \PM\ is updated using the sum of all events contributions.
See listing \ref{code} for pseudo-PyTorch code.

\begin{figure*}[tb]
\begin{lstlisting}[language=Python, caption={Pseudo-PyTorch Implementation of \algname}, label=code]
# event: Observed IC-event stream,  pose, motion: Initial estimates
def TeGRA(event, pose, motion, n_itr=100, eta=5e-4):
  for itr in range(max_itr):
    # Compute intensity at event pixel
    rgb_est = G(event.x, event.y, pose + event.u * motion)
    # Compute gradient of `rgb_est' w.r.t event timestamp
    gradt_est = torch.autograd.grad(rgb_est, event.u)  
    # Compute MSE between estimation and observation
    loss = mse_loss(gradt_est, event.r) 
    # Compute derivative w.r.t pose and motion
    loss.backward() 
    # Update pose and motion
    pose += eta*pose.grad; motion += eta*motion.grad
  return pose, motion
\end{lstlisting}
\end{figure*}
\section{Experiments}
\label{sec:experiments}

To demonstrate the effectiveness of the proposed \algname, we created the 6DOF camera pose tracking dataset for event data, called \dataname,  with ground-truth camera pose.
We used  BOP challenge 2020 scenes \cite{hodan2020bop} because it is photo-realistic, and the scene well represents the indoor localization scenario.
The BOP uses the BlenderProc \cite{denninger2019blenderproc} to render realistic images using ray tracing.
We first show the tracking result of \algname\ for proof of concept (POC)  (\secref{subsec:demo}).
Next, we show the intensive qualitative comparison with a dense algorithm (\figref{fig:rgb-based}) using intensity in terms of pose estimation accuracy (\secref{subsec:pose_est}).

\subsection{Event-based camera pose tracking dataset (EvTrack)}
\begin{figure*}[tb]
    \centering
    \includegraphics[width=1.0\linewidth]{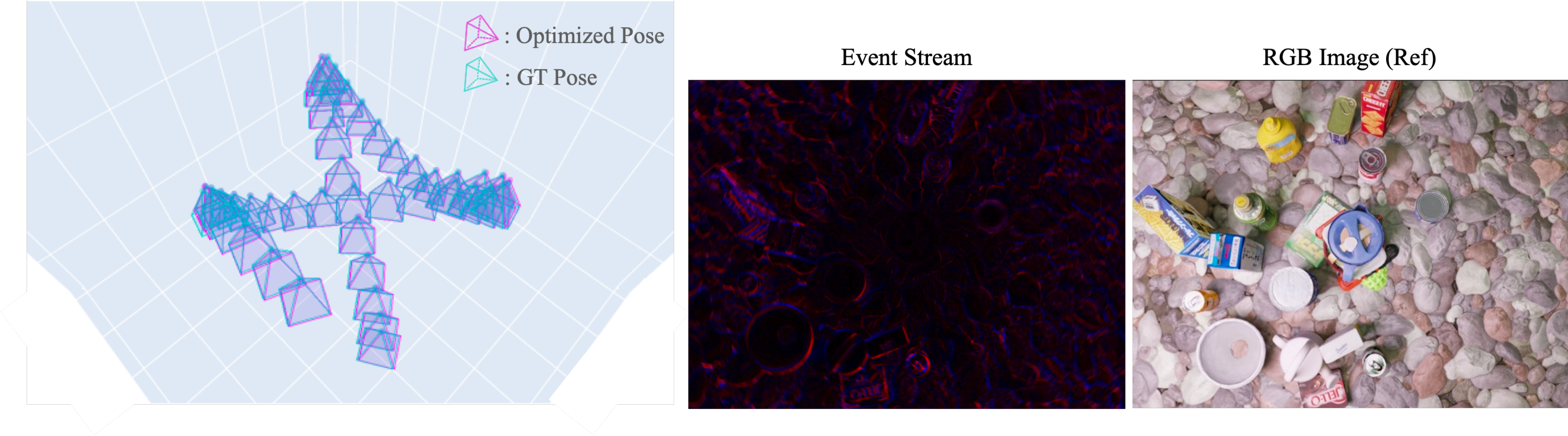}
    \caption{\small Camera pose tracking results from \textit{mix} scene (\secref{subsec:demo}); ground-truth ({\color{cyan}cyan}) and optimized trajectory ({\color{magenta}magenta}) (visualized in every 7 timestep).}
    \label{fig:globe}
\end{figure*}

The dataset consists of five scenes, \textit{mix}, \textit{hb}, \textit{lm}, \textit{tyol} and \textit{ycbb}, the last four scenes correspond to the BOP data split and \textit{mix} scenes include all the four data to simulate the ordinary indoor situation.
We used \textit{mix} scene for proof of concept in the tracking scenario and use the other four for the quantitative evaluation in terms of pose estimation accuracy.
We generated 500 images for \nerf\ training and three camera trajectories (\figref{fig:globe} right) for each of the five scenes simulating the drone hovering around a room.
The IC-event stream is generated by using pairs of consecutive images (total of 1,000), $\{L(t), L(t+\nu)\}$ ($L(t) \in \mathbb{R}^{H\times W}$), by subtracting them.
The size of image $(H,W)$ is $(480\times 640)$.
The threshold $\delta$ for triggering the event is set to $0.05$ (intensities are normalized to $[0,1]$) for all scenes.
In our experiment, we converted the RGB IC event to a grayscale IC event since most event cameras detect grayscale intensity changes.
Each event stream is generated using five consecutive frames, assuming the motion is approximately linear within the interval. 

\begin{table*}[tb]
\centering
\caption{Pose estimation accuracy comparison of existing dense approach (\figref{fig:rgb-based}) and our sparse approach (\figref{fig:tegn}, \algname). 
}
\vspace{5pt}
\scalebox{1.0}{
\begin{tabular}{cc||cc|cc|cc|cc}
\hline
                                                  &      & \multicolumn{2}{c|}{\textit{hb}} & \multicolumn{2}{c|}{\textit{lm}} & \multicolumn{2}{c|}{\textit{tyol}} & \multicolumn{2}{c}{\textit{ycbv}} \\
                                                  &      & Rot.      & Trans.     & Rot.      & Trans.     & Rot.       & Trans.      & Rot.       & Trans. \\ \hline
\multirow{3}{*}{Dense}                            & seq0 &0.034      &0.0002      &0.028      &0.0004      &0.034       &0.0003       &0.028       &0.0003       \\
                                                  & seq1 &0.036      &0.0002      &0.562      &0.0003      &0.033       &0.0002       &0.567       &0.0004       \\
                                                  & seq2 &0.030      &0.0002      &0.273      &0.0004      &0.028       &0.0002       &0.269       &0.0004       \\ \hline
\multirow{3}{*}{Sparse (ours)}                    & seq0 &0.217      &0.0018      &0.321      &0.0047      &0.350       &0.0030       &0.760       &0.0021             \\
                                                  & seq1 &0.448      &0.0065      &0.790      &0.0102      &0.477       &0.0053       &0.448       &0.0072       \\
                                                  & seq2 &0.338      &0.0054      &0.362      &0.0088      &0.425       &0.0058       &0.290       &0.0051        \\ \hline
\end{tabular}}
\label{tab:result}
\end{table*}

\subsection{Implementation}
We use Pytorch \cite{NEURIPS2019_9015} to train the \nerf-model and run \algname\ for tracking\footnotemark. 

\noindent\textbf{Training}
We follow the network settings from the original \nerf\ with minor modifications; use the $\texttt{softplus}$ activation for the volume density $\sigma$ as recommended in BARF \cite{lin2021barf} for improved stability.

\noindent\textbf{Tracking}
We add the RGB-to-gray layer (\secref{sec:tracking_alg}) to use grayscale IC-events.
We randomly select 750 event pixels from the observed event and update the \PM\ for $\texttt{n\_itr}$ (1,000) times; it amounts to  \invsparcity\ of the dense algorithm.
The learning rate $\eta$ of the $S, \dot{S}$ is set to $5 \times 10^{-5}$ exponentially decaying to $5 \times 10^{-6}$ toward $\texttt{n\_itr}$.
The temporal gradient of intensity w.r.t the event timestamp $e_u$ is computed by using Pytorch's $\texttt{autograd.grad}$.

\subsection{Proof of Concept (Tracking)}
\label{subsec:POC}
\footnotetext{See supplement for network architecture of \nerf\ (supplement-\ref{suppsec:network}), and frame-based dense tracking algorithm which we compare the accuracy (supplement-\ref{suppsec:dense_track}).} 
\label{subsec:demo}
For the POC of the proposed idea, we applied \algname\ for tracking.
We used \textit{seq0} from \textit{mix} scene.
We use estimated \PM\ from the previous timestep to initialize $(S^{\mathrm{ini}}_t,\dot{S}^{\mathrm{ini}}_t)$ in the next timestep as discussed in \secref{sec:tracking_alg}.
The results are shown in \figref{fig:globe}.
We confirmed that the pose is successfully tracked without drifting\footnote{Video for this tracking is included in the supplement. See supplement-\ref{suppsec:add_exp} for additional results on \nerf\ dataset.}.
Finally, the pose estimation error was $(0.13\degree, 0.0003)$.
The average number of events per pose update was \invsparcity\ of the entire pixel.

\subsection{Quantitave Benchmarks (Pose Estimation)}
\label{subsec:pose_est}
To quantitatively evaluate the performance of the \algname, we compare the pose estimation accuracy with the dense algorithm using the difference in intensity (\figref{fig:rgb-based}).
In this experiment, we randomly initialized the pose and motion for each stream.
The results are shown in \tabref{tab:result}.
Both achieved comparable accuracy in maintaining camera-pose tracking, while ours used only 2.4\% of pixels.

\section{Conclusion}
How can we utilize sparse events for recovering the camera pose?
Answer: Camera pose is recovered by minimizing the error between the temporal gradient of the scene represented as a \nerf\ and sparse events. 
Our tracking algorithm, \algname, could update the pose using sparse event points.
This mechanism is a significant advantage over the existing image-space algorithm, which requires dense computation.
We demonstrate \algname\ in a tracking scenario with unseen background clutter.

We believe the proposed idea opens the door for realizing an event-based camera pose tracking using implicit 3D-scene representation.
This study focuses on demonstrating the algorithm in a naive setup; therefore, we left large areas for future work, either from an experimental or algorithmic perspective.

\subsection{Application to Real-World Data}
\label{sssec:real_world}
One of our ultimate goals is to utilize the proposed method in the practical autonomous driving scenario.
More specifically, as future work, we plan to apply the \algname\ for Block-NeRF \cite{tancik2022blocknerf}, which realized scaling up the \nerf\ into a city-scale automotive environment.
The \algname\ could incorporate the mip-NeRF \cite{barron2021mipnerf} rendering algorithm, which is key to realizing the large-scale modeling in Block-NeRF.
The principle we present (theorem-\ref{prop}) is compatible with a variety of \nerf-variant as long as the 3D-scene is represented in the form of \myeqref{eq:nerf}, where $G_{\mathcal{M}}$ is differentiable w.r.t pose $S$.

\subsection{Asynchronous Update using Binary-Event}
\label{sssec:async_update}
In this study, we use IC-event instead of binary-event, and synchronous event stream instead of asynchronous one.
We choose this experimental setup mainly due to the implementational difficulties in generating event streams.
It requires engineering effort to generate asynchronous binary-event streams, such as modifying event-camera simulators like ESIM \cite{rebecq2018esim}.
To make the algorithm compatible with the binary polarity $e_p$, the loss term in \myeqref{eq:theorem} needs to be modified slightly:
\begin{equation}
\label{eq:loss_bin}
    \mathcal{L}_{bin} = {\sum_{i}\left\|\operatorname{SoftSgn}\left ( \frac{\partial G_{\mathcal{M}}\left(e_x^i, e_y^i, S_t+\bar{e}_u^i\dot{S}_t\right)}{\partial \bar{e}_u^i}\right )  - e_p^i\right\|_2},
\end{equation}
where the $\operatorname{SoftSgn}$ is a soft version of the sign function, which maps continuous intensity change into (soft) polarity.
We left this exploration as future work.

\subsection{Speed Up}
\label{sssec:speed_up}
Thanks to the sparse mechanism of \algname, the number of pixels to be evaluated for computing the pose update is significantly lower than the entire pixel (\secref{subsec:demo}).
Speeding up the \nerf\ for real-time rendering is an active research topic \cite{garbin2021fastnerf,lindell2021autoint,yu2021plenoctrees,yu_and_fridovichkeil2021plenoxels}.
For example, FastNeRF \cite{garbin2021fastnerf} utilizes a separate network for position-dependent MLP and direction-dependent MLP for speed up.
As discussed above, the proposed mechanism is compatible with other \nerf-variants.
We expect combining our sparse mechanism with these approaches is a vital topic to realize real-time tracking on mobile devices. 
We'll incorporate the method and then evaluate the FLOPS and wall-clock time.

\subsection{Extention to SLAM}
\label{sssec:slam}
It is an exciting research direction to extend our algorithm into simultaneous localization and mapping (SLAM).
Now, \nerf\ is emerging as an entirely new framework for SLAM \cite{sucar2021imap,lin2021barf,wang2021nerf,zhu2021nice}.
iMAP \cite{sucar2021imap} is pioneering work utilizing \nerf\ for realizing real-time SLAM.
We expect incorporating \algname\ will significantly speed up the \nerf-based SLAM.

{\small
\bibliographystyle{ieee_fullname}
\bibliography{egbib}
}
\newpage
\clearpage

\renewcommand{\thesection}{\Alph{section}}
\renewcommand{\thetable}{\Alph{section}}
\renewcommand{\thefigure}{\Alph{section}}
\renewcommand{\thealgorithm}{\Alph{section}}
\setcounter{section}{0}
\setcounter{figure}{0}
\setcounter{table}{0}
\setcounter{equation}{0}

\section{NeRF}
\subsection{NeRF Formulation}
\nerf\ \cite{mildenhall2020nerf} is an implicit 3D scene representation using neural networks, originally proposed for novel view synthesis.
As in \secref{sec:preliminaries}, \nerf, $G_{\mathcal{M}}$, takes an image coordinate $(x, y)$ and 6DoF pose $S\in \se (3)$ of the camera as inputs and render the RGB intensity $\mathbf{c}$ at that coordinate.
\nerf\ encodes a 3D scene as a continuous representation using an MLP $f_\mathcal{M}$, parameterized by learned parameter $\mathcal{M}$.
The $f_\mathcal{M}$ output the intensity $\mathbf{c}'$ and volume density $\sigma$ given  a viewing ray $\mathbf{d}$ and a 3D coordinate $\mathbf{x}$.
The viewing ray $\mathbf{d}$ is computed from the camera pose $S$ and image coordinate $(x, y)$.
Let the homogeneous image coordinate $\bar{\mathbf{u}}=(x, y, 1)$; then, the 3D point $\mathbf{x}^i$ along the viewing ray $\mathbf{d}$ at depth $z^i$ is expressed as $\mathbf{x}^i = \bar{\mathbf{u}} + z^i\mathbf{d}$.
\nerf\ integrate the intensity $\mathbf{c}' $ using  density $\sigma$ by volume rendering to obtain intensity at the image coordinate $(x,y)$ for the given camera pose $S$ as follows:
\begin{align}
\label{eq:G}
    G_\mathcal{M}(x, y, S)&=\int_{z_{n}}^{z_{f}} T(z) \sigma(\bar{\mathbf{u}}+z\mathbf{d}) \mathbf{c}'(\bar{\mathbf{u}}+z\mathbf{d}) dz,\\
    T(z)&=\exp \left(-\int_{z_{n}}^{z} \sigma(\bar{\mathbf{u}}+z'\mathbf{d}) dz'\right)
\end{align}
where, $z_n$ and $z_f$ are bounds on the depth range of interest.
In practice, this rendering function is approximated numerically via quadrature on points in the depth direction.

\subsection{Network Architecture}
\label{suppsec:network}
\begin{figure}[ht]
    \centering
    \includegraphics[width=1.0\linewidth]{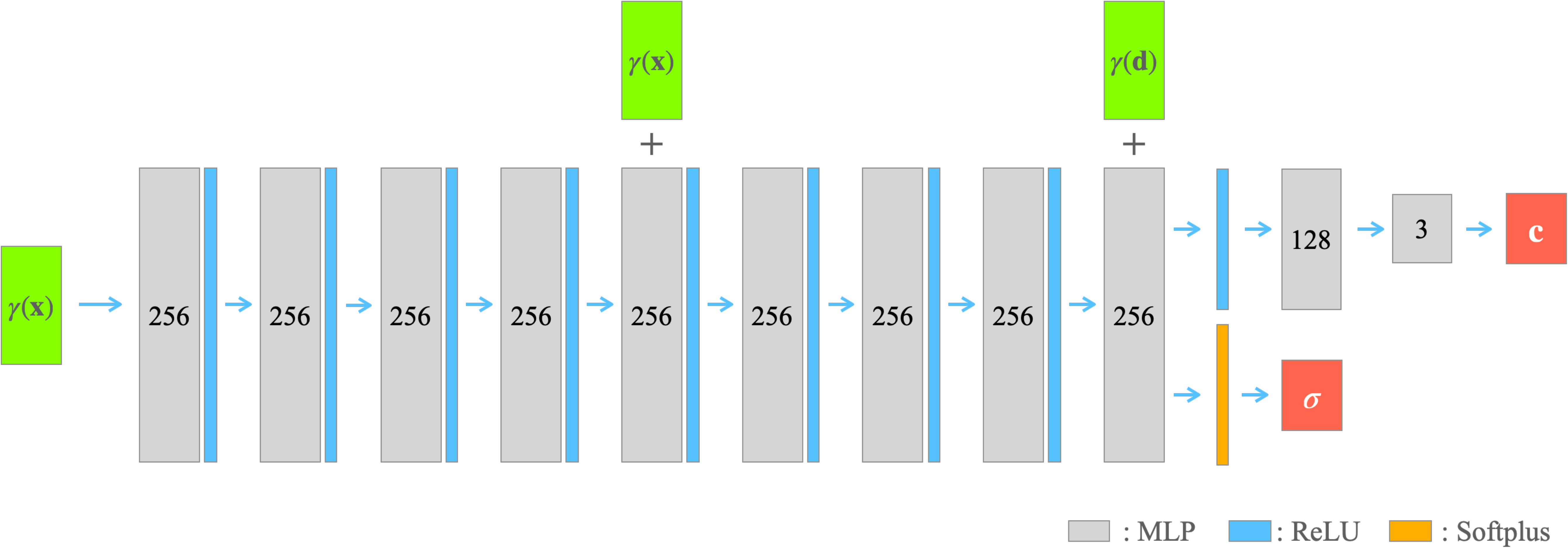}
    \caption{The \nerf\ network architecture used for the experiments.}
    \label{fig:network}
\end{figure}

The \nerf\ network architecture used for the experiments is shown in \figref{fig:network}.
We adopt code of \nerf\ from BARF official implementation \url{https://github.com/chenhsuanlin/bundle-adjusting-NeRF} for training the model. 
The difference from the original \nerf\ \cite{mildenhall2020nerf} is activation for density $\sigma$ is replaced by \texttt{softplus} from \texttt{relu} to improve stability during training. The function $\gamma$ is a positional encoding function that maps inputs 3D coordinates  $x$ to higher dimensions of different sinusoidal frequency basis functions.

\newpage
\section{Event-based camera pose tracking dataset (EvTrack)}
\label{suppsec:EvTrack}
We created the 6DOF camera pose tracking dataset for event data, called \dataname,  with ground-truth camera pose to demonstrate the effectiveness of the proposed \algname\ and foster subsequent studies of event-based camera pose tracking using implicit 3D scene representation.

We used  BOP challenge 2020 scenes \cite{hodan2020bop} because it is photo-realistic, and the scene represents the indoor localization scenario well.
The BOP 2020 uses the BlenderProc \cite{denninger2019blenderproc} to render realistic images using ray-tracing.
The dataset consists of five scenes, \textit{mix}, \textit{hb}, \textit{lm}, \textit{tyol} and \textit{ycbb}, the last four scenes correspond to the BOP data split and \textit{mix} scenes include all the four data to simulate the ordinary indoor situation.

\subsection{Data for training}
The training data consists of 500 images  with ground-truth camera poses for each scene.
Each camera pose is randomly sampled from a pre-defined rectangular space that includes the tracking trajectories described in \ref{suppsec:tracking_data}.
The size of image $(H,W)$ is $(480\times 640)$.
An implicit 3D scene representation such as \nerf\ can be trained using the data.

\subsection{Data for testing (tracking)}
\label{suppsec:tracking_data}
The test data consists of three camera trajectories for each scene  simulating the drone hovering around a room (\figref{fig:globe} right).
For each scene, we render 1,000 consecutive images with ground-truth camera poses.
The size of image $(H,W)$ is $(480\times 640)$.
Using the consecutive images, we generate IC-event by subtracting the successive frames.
The threshold $\delta$ for triggering the event is set to $0.05$ (intensities are normalized to $[0,1]$) for all scenes.

In our experiment, we use grayscale IC-event since most event cameras detect grayscale intensity changes.
To generate a grayscale IC-event,  we convert the RGB frame into a grayscale frame using  $\texttt{OpenCV}$'s  $\texttt{cvtColor()}$ function.

\subsection{Chunk size}
The number of events consumed by the network at once (chunk size) must be determined depending on the scene.
In general, we can expect a higher accuracy when using a longer chunk size as long as the chunked event stream conforms to the  motion model (we used the linear model of \eqref{eq:theorem} in our experiment).
On the other hand, when the chunk size is too large, and the chunked event stream contains non-linear motion, the observed event can not be explained by the motion model, and the performance will degrade.
We use \textit{four} timestep (generated from five consecutive frames) in our main experiment.
The camera motion in the \dataname\  can be considered  approximately linear within the \textit{five} frame interval. 

\subsection{Required accuracy for tracking}
In the main paper, we discussed both dense approach and our sparse approach to achieve comparable accuracy to keep the tracking (although their pose error differs).
The pose error from \algname\ is larger than the dense approach because the observed event trajectory does not precisely conform to the motion model we use (linear motion model), while the dense model aligns a single image without temporal information (therefore, there is no model error).
One could utilize a non-linear motion model to reduce the error further. 
However, the error from \algname\ is  sufficiently small to keep stable tracking; therefore, the linear model suffices for tracking (on the \dataname\ dataset).
\tabref{tab:result_full} show the tracking results on all the sequences in the \dataname.
We consider the pose is successfully tracked ($\checkmark$) if the error in the final pose is less than the initial pose error of $(1.0\degree, 0.001)$ (Due to the time required to run the dense RGB method (\figref{fig:rgb-based}), we used the randomly sampled 50 consecutive frames from each sequence).

We also confirmed that \algname\ could successfully track the camera pose without drifting on the \nerf's blender scene (supplement-\ref{suppsec:add_exp}). 

\begin{table}[ht]
\centering
\caption{Tracking test of dense approach (\figref{fig:rgb-based}) and our sparse approach (\figref{fig:tegn}, \algname). 
}
\vspace{5pt}
\scalebox{0.8}{
\begin{tabular}{cc||ccccc}
\hline
                                                  &      & \textit{mix} & \textit{hb} & \textit{lm} & \textit{tyol} & \textit{ycbv} \\ \hline
\multirow{3}{*}{Dense}                            & seq0 & $\checkmark$ &$\checkmark$ &$\checkmark$ &$\checkmark$   &$\checkmark$   \\
                                                  & seq1 & $\checkmark$ &$\checkmark$ &$\checkmark$ &$\checkmark$   &$\checkmark$   \\
                                                  & seq2 &$\checkmark$  &$\checkmark$ &$\checkmark$ &$\checkmark$   &$\checkmark$   \\ \hline
\multirow{3}{*}{Sparse (ours)}                    & seq0 &$\checkmark$  &$\checkmark$ &$\checkmark$ &$\checkmark$   &$\checkmark$   \\
                                                  & seq1 &$\checkmark$  &$\checkmark$ &$\checkmark$ &$\checkmark$   &$\checkmark$   \\
                                                  & seq2 &$\checkmark$  &$\checkmark$ &$\checkmark$ &$\checkmark$   &$\checkmark$   \\ \hline
\end{tabular}}
\label{tab:result_full}
\end{table}

\section{Tracking using dense image}
\label{suppsec:dense_track}
As discussed in the \secref{sec:intro} and (\figref{fig:rgb-based}), camera pose can be recovered using \nerf\ and dense intensity observation.
We use the same pre-trained \nerf\ model as our \algname\ and estimate the camera pose $S$.
Unlike ours, which optimizes pose $S$ and motion $\dot{S}$, the frame-based algorithm only optimizes pose $S$.
Because the entire $480\times 640$ pixels can not fit into the single GPU, we split the pixels into a 4,000-pixel chunk and accumulated the gradient w.r.t. $S$ to compute the single pose update for a frame.
The pose update is repeated $\operatorname{n\_itr}$ (100) times for each observed frame.
The learning rate $\eta$ of the $S$ is set to $5 \times 10^{-4}$ exponentially decaying to $5 \times 10^{-5}$ toward $\operatorname{n\_itr}$.

\newpage
\section{Binary Event to IC-Event}
\label{suppsec:b2e}
In this study, we use IC-event instead of binary-event, and synchronous event stream instead of asynchronous one.
We choose this experimental setup mainly due to the implementational difficulties in generating event streams.
It requires engineering effort to generate asynchronous binary-event streams, such as modifying event-camera simulators.
Some event-based cameras can detect the intensity change $e_r$ directly, such as Celex-V \cite{chen2019live}.
However, perhaps the most common type reports an event $e=[e_x, e_y, e_t, e_p]^\mathsf{T}$ when it detects the intensity change $\Delta L$ by a specified amount $\delta$, where $e_p\in[+1, -1]$ is the polarity which indicates the increase or decrease of the intensity changes.

We consider there are two possible approaches to make \algname\ compatible with the binary-event; 1) Modify the loss function of \eqref{eq:theorem} to adapt to the binary-event (\secref{sssec:async_update}), 2) Convert the binary-event to IC-event using the timestamp (supplement-\ref{suppsec:b2e}). In this work, we use IC-event for simplicity and left the exploration for future work.

\begin{figure}[ht]
    \centering
    \includegraphics[width=1.0\linewidth]{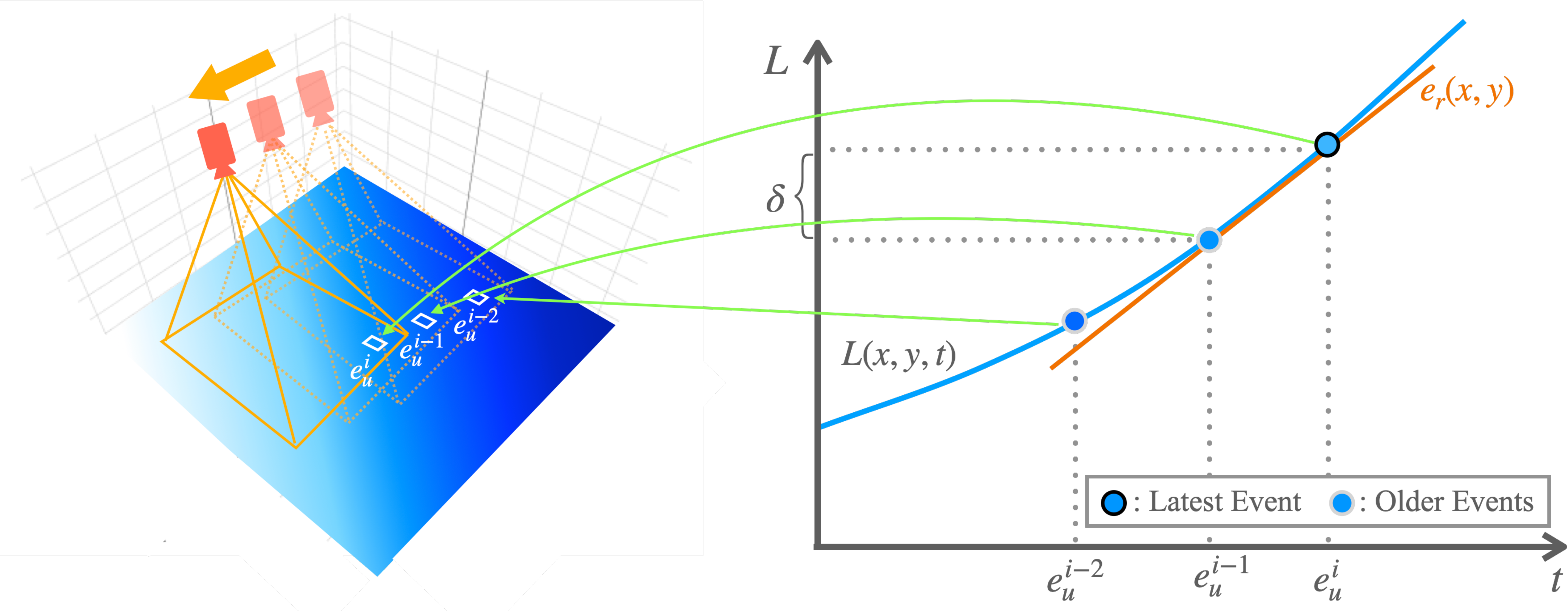}
    \caption{Binary-Event to IC-Event}
    \label{fig:graphs}
\end{figure}

In this section, we discuss the second approach.
IC-event could be estimated by a simple (linear fitting).
One can use the time difference between the latest two consecutive events on each pixel to estimate the intensity difference, as depicted in \figref{fig:graphs}.
There are other options, such as fitting higher-order functions using  more events.

\newpage

\section{Algorithm of TeGRA}
\label{suppsec:alg}

The complete algorithm of \algname\ is listed in Alg.\ref{alg}.
Note that the algorithm listed in Alg.\ref{alg} is slightly different from the one we used in our main experiment (listing-\ref{code}).
In the main experiments, we used a fixed number of iterations for a fair comparison; however, we expect  terminating iteration using the norm of the \PM\ update is preferable for the practical application in terms of computational efficiency.

\renewcommand{\algorithmicrequire}{\textbf{Input:}}
\renewcommand{\algorithmicensure}{\textbf{Output:}}
\begin{algorithm}
\footnotesize
    \caption{\algname}
    \label{alg1}
    \begin{algorithmic}[1]
    \REQUIRE IC-event stream $\mathbf{e}_t$\\
             Optimized \PM\ $(S^{\mathrm{opt}}_{t-\nu}, \dot{S}^{\mathrm{opt}}_{t-\nu})$ at time ${t-\nu}$
    \STATE $\epsilon \leftarrow \inf$
    \STATE $({S}_t^{\mathrm{ini}}, \dot{S}^{\mathrm{ini}}_t) \leftarrow (S^{opt}_{t-\nu}+\nu \dot{S}^{\mathrm{opt}}_{t-\nu}, \dot{S}^{\mathrm{opt}}_{t-\nu})$ \hfill $\triangleright{(\textrm{linear 
 motion})}$
    \STATE $({S}, {\dot{S}}) \leftarrow (S^{\mathrm{ini}}_t, \dot{S}^{\mathrm{ini}}_t)$
    \WHILE {$\epsilon > \epsilon_{thr}$}
    \STATE Feedward event coordinate and pose to $G_{\mathcal{M}}$
    \STATE Compute temporal gradient of $G_{\mathcal{M}}$ w.r.t $[...,\bar{e}_u^i,...]$
    \STATE Evaluate error between \modelname\ and IC-event \hfill $\triangleright{\text{\myeqref{eq:theorem}}}$
    \STATE Backpropagete the error thruough \modelname\
    \STATE Update pose \& motion
    \STATE $\epsilon \leftarrow$ norm of \PM\ update
    \ENDWHILE
    \STATE $({S}_t^{\mathrm{opt}}, \dot{S}^{\mathrm{opt}}_t) \leftarrow (S, \dot{S})$
    \ENSURE $({S}_t^{\mathrm{opt}}, \dot{S}^{\mathrm{opt}}_t)$
    \end{algorithmic}
    \label{alg}
\end{algorithm}

\newpage
\section{Additional tracking result on  NeRF dataset}
\label{suppsec:add_exp}
\begin{figure}[ht]
    \centering
    \includegraphics[width=1.0\linewidth]{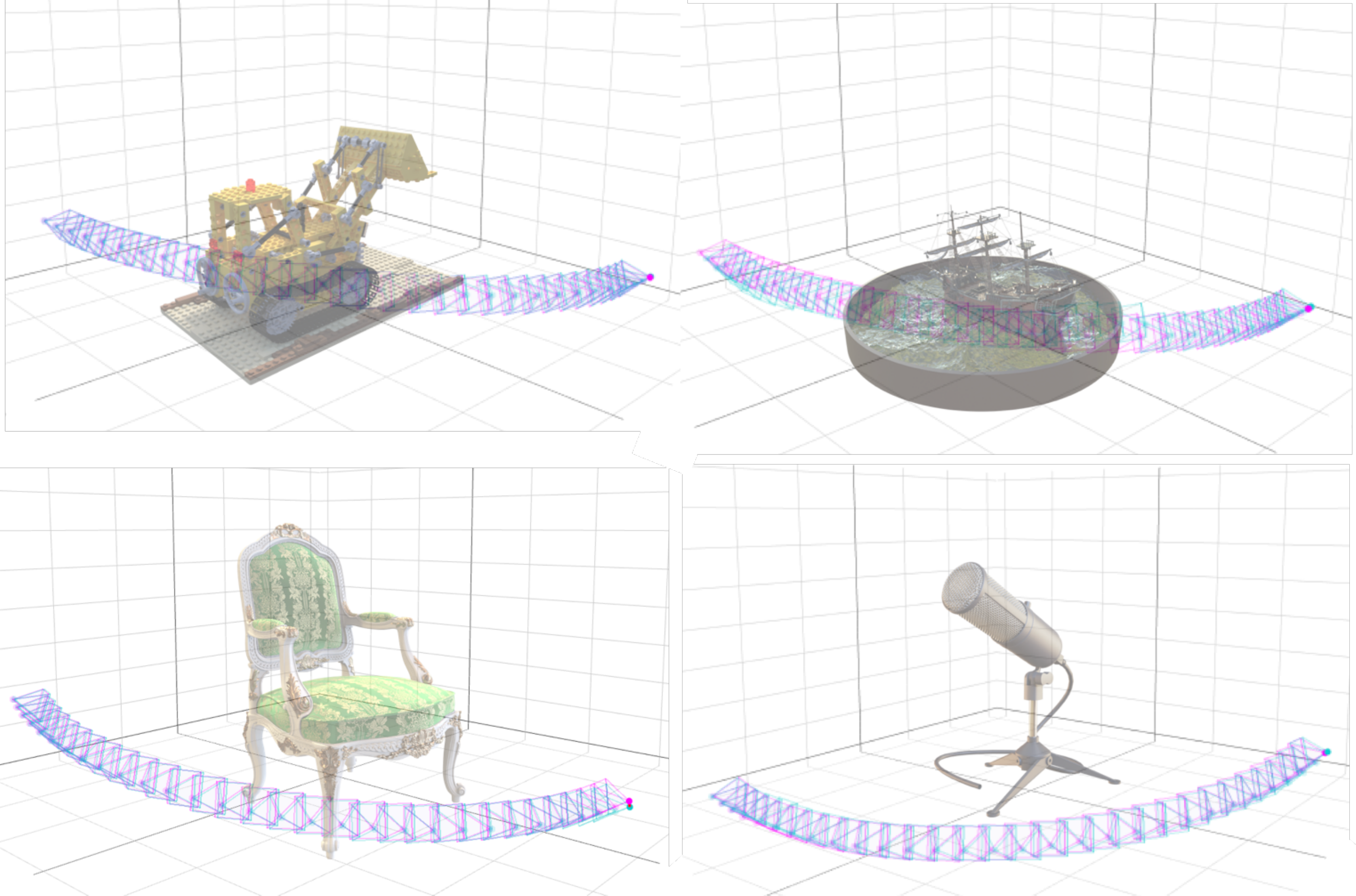}
    \vspace{3mm}
    \caption{Additional tracking results on \nerf's blender scene.}
    \label{fig:additional_tracking_result}
\end{figure}

In addition to the main tracking experiment using \dataname, we also conducted an  experiment to evaluate the tracking performance of \algname\ on the \nerf's blender scene \cite{mildenhall2020nerf}.
For this experiment, we created the event stream from 1,000 consecutive frames using the blender scene; the camera rotates around the object while moving up and down.
We also confirmed that the camera pose is successfully tracked without drift on this dataset (\figref{fig:additional_tracking_result}, See also supplemental video).

\end{document}